\documentclass[11pt]{article}

\usepackage[final]{acl}

\usepackage{times}
\usepackage{latexsym}

\usepackage[T1]{fontenc}

\usepackage[utf8]{inputenc}

\usepackage{microtype}
\usepackage{enumitem}
\usepackage{booktabs}
\usepackage{multirow}
\usepackage{tabularray}
\usepackage{makecell}
\usepackage{inconsolata}
\usepackage{amsmath}
\usepackage{xcolor}
\usepackage{colortbl}
\usepackage{graphicx}
\usepackage{enumitem}
\usepackage{amssymb}
\usepackage{color}

\usepackage{pifont}
\newcommand{\cmark}{\ding{51}}
\newcommand{\xmark}{\ding{55}}

\newlength\savewidth

\title{Urban-R1: Reinforced MLLMs Mitigate Geospatial Biases \\ for Urban General Intelligence}

\author{
 \textbf{Qiongyan Wang\textsuperscript{1}},
 \textbf{Xingchen Zou\textsuperscript{1}},
 \textbf{Yutian Jiang\textsuperscript{1}},
 \textbf{Haomin Wen\textsuperscript{2}},
\\
 \textbf{Jiaheng Wei\textsuperscript{1}},
 \textbf{Qingsong Wen\textsuperscript{3}},
 \textbf{Yuxuan Liang\textsuperscript{1,*}}
\\
 \textsuperscript{1}The Hong Kong University of Science and Technology (Guangzhou) \\
 \textsuperscript{2}Carnegie Mellon University
 \textsuperscript{3}Squirrel Ai Learning
\\
 \small{
   \textbf{Correspondence:} \href{mailto:email@domain}{yuxliang@outlook.com}
 }
}

\begin{document}
\maketitle
\begin{abstract}

Rapid urbanization intensifies the demand for Urban General Intelligence (UGI), referring to AI systems that can understand and reason about complex urban environments. Recent studies have built urban foundation models using supervised fine-tuning (SFT) of LLMs and MLLMs, yet these models exhibit persistent geospatial bias, producing regionally skewed predictions and limited generalization. To this end, we propose Urban-R1, a reinforcement learning–based post-training framework that aligns MLLMs with the objectives of UGI. Urban-R1 adopts Group Relative Policy Optimization (GRPO) to optimize reasoning across geographic groups and employs urban region profiling as a proxy task to provide measurable rewards from multimodal urban data. Extensive experiments across diverse regions and tasks show that Urban-R1 effectively mitigates geo-bias and improves cross-region generalization, outperforming both SFT-trained and closed-source models. Our results highlight reinforcement learning alignment as a promising pathway toward equitable and trustworthy urban intelligence.
\end{abstract}

\section{Introduction}

Rapid urbanization is reshaping paradigms of city planning and management, intensifying the demand for \textbf{Urban General Intelligence (UGI)}, i.e., advanced AI systems capable of understanding, interpreting, and managing complex urban environments~\cite{zhang2024towards,liang2025foundation}. UGI aspires to move beyond traditional task-specific models (e.g., traffic forecasting) toward general-purpose agents that autonomously handle diverse urban challenges such as GDP estimation and urban planning. Achieving this vision requires models that can effectively interpret multimodal urban data (e.g., satellite imagery, geo-coordinates) and provide adaptive decision-making across heterogeneous real-world scenarios.

\begin{figure}[!t]
    \centering
    \includegraphics[width=1\linewidth]{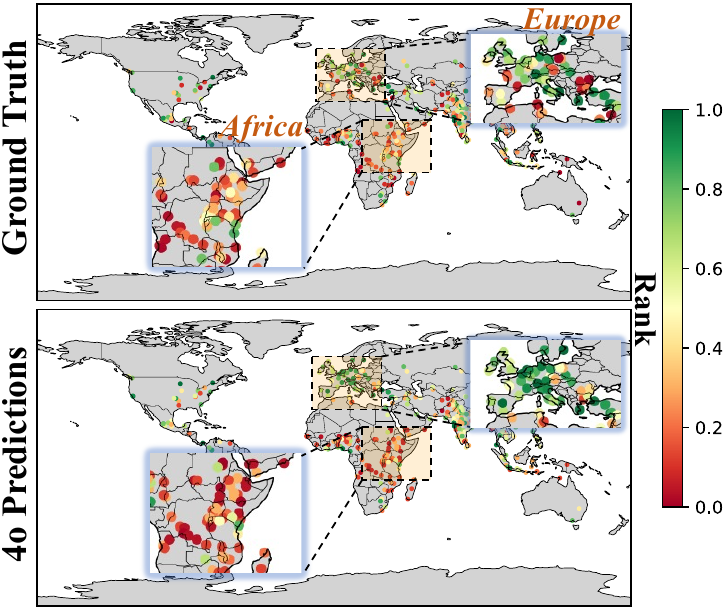}
    \caption{An example of geographic bias in regional GDP prediction by GPT-4o. The top panel depicts the ground truth of GDP rankings (color-coded: green indicates a higher rank, red indicates a lower rank), while the bottom panel shows predictions from GPT-4o. }
    
    \label{fig:intro}
    \vspace{-1em}
\end{figure}

To advance UGI, recent studies have leveraged Large Language Models (LLMs) and Multimodal LLMs (MLLMs) to construct urban foundation models through {Supervised Fine-Tuning (SFT)}. For instance, GeoChat employs lightweight LoRA adaptation for remote-sensing imagery \cite{kuckreja2024geochat}, CityGPT fine-tunes LLMs on structured geospatial data \cite{feng2025citygpt}, and UrbanLLaVA extends vision–language models to urban imagery understanding \cite{feng2025urbanllava}. These efforts demonstrate the promise of language-centric or multimodal backbones for unifying heterogeneous urban inputs, implicitly encoding non-trivial geospatial knowledge, and achieving strong in-domain performance when trained and evaluated on similar data distributions.

Though promising, urban foundation models built on SFT still face a fundamental challenge -- \textbf{Geospatial Bias (geo-bias)} \cite{manvi2024large}, which refers to systematic deviations between model predictions and real-world geographic distributions. As shown in Figure \ref{fig:intro}, even a powerful model like GPT-4o tends to overestimate GDP in European regions while underestimating it in African regions. Such bias arises not only from data imbalance but also from the model’s limited ability to adapt to new or underrepresented spatial contexts. In practice, this leads to severe consequences: when deployed in real-world urban systems, geo-biased models may produce skewed policy recommendations, misestimate regional development, or unfairly allocate resources across different geographies. Therefore, mitigating geo-bias is essential for achieving trustworthy and equitable UGI.

From a mechanism perspective, geo-bias is deeply rooted in the training paradigm of SFT. By minimizing token-level prediction errors, SFT encourages models to replicate the conditional distribution of their training data. This process makes the model highly dependent on empirical correlations rather than invariant geographic relationships. As a result, the model learns surface patterns, such as ``dense infrastructure implies high income'', that hold in dominant regions but break down elsewhere. When applied to unseen or underrepresented areas, the model’s reasoning collapses to these spurious correlations, producing systematic geographic distortions. In short, \emph{SFT optimizes for pattern imitation rather than causal generalization}, which inevitably limits its ability to reason about cities beyond the scope of the training set.

To overcome these issues, we introduce \textbf{Urban-R1}, a reinforcement learning (RL)–based post-training framework that aligns MLLMs with the objectives of UGI. RL offers a fundamentally different optimization paradigm: instead of imitating human-labeled answers, the model learns by \emph{maximizing explicit reward signals that evaluate the rationality, accuracy, and consistency of its geographic reasoning}. In particular, Urban-R1 adopts Group Relative Policy Optimization (GRPO), which compares multiple reasoning trajectories within the same geographic group and updates the policy to favor those reflecting invariant and evidence-grounded spatial relations. This intra-group optimization allows the model to learn reasoning patterns that are robust across regions and less dependent on biased training distributions. Furthermore, we propose \textbf{Urban Region Profiling (URP)} as a proxy task for RL alignment. URP integrates multimodal urban data (e.g., satellite imagery, coordinates, and textual context) to estimate objective indicators such as GDP, population, and carbon emissions. These indicators provide measurable and verifiable rewards that guide the model toward stable and transferable geography-aware reasoning.

In summary, our contributions are threefold:
\begin{itemize}[leftmargin=*,topsep=0em,itemsep=0em]
    \item \textbf{Urban-R1: A paradigm shift for urban general intelligence.}
We move beyond supervised imitation toward reinforcement-based alignment, demonstrating that reinforcement learning can serve as an effective mechanism to endow multimodal models with geography-aware reasoning and reduce systemic geospatial biases.
    \item \textbf{A task formulation bridging learning and evaluation.} We cast Urban Region Profiling (URP) as a principled proxy for urban reasoning, providing a measurable, transferable setting that connects multimodal perception with quantitative urban understanding. This formulation enables reward-driven optimization of reasoning quality without reliance on dense supervision.

    \item \textbf{Extensive experiments.} Through comprehensive evaluation across diverse regions and urban tasks, our Urban-R1 shows that RL-based models not only outperform SFT baselines but also yield more accurate results against closed-source LLMs/MLLMs in multiple urban reasoning tasks, marking a promising direction for equitable and trustworthy urban AI.
\end{itemize}

\section{Related Works}
\subsection{Urban General Intelligence}
Urban General Intelligence has evolved along multiple reinforcing paths.
Early work used \emph{sparse signals} (e.g., nighttime lights or single-source satellite imagery) as proxies for wealth, economic activity, or housing outcomes. These task- and city-specific predictors target narrow outcomes, generalize poorly across regions, and require dedicated labels \cite{yeh2020using,park2022learning,he2018perceiving,huang2021m3g,law2019take}.
To improve transferability and reduce annotation costs, research moved toward \emph{unified multimodal representations} that combine imagery with spatial and textual context through self-supervised objectives, yielding better transfer than task-specific pipelines \cite{jean2019tile2vec,wang2020urban2vec,bjorck2021accelerating,kang2020deep,xi2022beyond}. More recently, vision--language \emph{contrastive pretraining} has advanced state-of-the-art region representations by injecting textual semantics \cite{yan2024urbanclip,hao2024urbanvlp}. Even so, such representations generally still require \emph{per-task} fine-tuning for downstream objectives.

To address this limitation, recent work leverages large models from general areas with instruction-based \emph{supervised fine-tuning} to strengthen models' internal understanding of urban patterns (e.g., spatio-temporal reasoning and domain grounding), improving in-distribution performance on profiling queries \cite{li2024urbangpt,xiao2024refound,lai2025llmlight}. However, evaluations of LLM-based pipelines reveal limited cross-task and cross-city transfer, often exhibiting geographic bias~\cite{zou2025traffic,liu2025reinforcement,cao2024survey} and weak visual grounding, which suggests that supervised fine-tuning alone does not suffice for reliable urban reasoning~\cite{manvi2023geollm,manvi2024large}.


\subsection{Reinforcement Learning for Large Models}
Reinforcement learning (RL) fine-tuning provides an alternative to instruction fine-tuning, but classic RLHF pipelines have seen limited adoption in practice because of training complexity and compute cost~\cite{ouyang2022training,gao2023scaling,bai2022training,ramamurthy2022reinforcement}. DeepSeek-R1~\cite{liu2024deepseek} validates GRPO by showing that RL alignment can improve LLMs' understanding with limited training data and lightweight reward computation; importantly, RL trains models to reason about problems and produce solutions rather than merely memorizing answers~\cite{liu2025reinforcement,chu2025sft}. Building on this, several works adapted GRPO to urban settings: Traffic-R1 applies GRPO to traffic-signal control and reports domain-level gains through task- specific RL~\cite{zou2025traffic}, and Vision-R1~\cite{huang2025vision} extends GRPO to Multimodal Large Language Models (MLLMs), improving multimodal understanding by optimizing accuracy and format/parsability rewards. Nonetheless, RL alignment of large models for broad urban-region understanding and for general improvement across urban knowledge tasks with less geospatial biases remains underexplored, which motivates our RL-based approach.

\section{Methodology}

\begin{figure}[!t]
    \centering
    \includegraphics[width=1\linewidth]{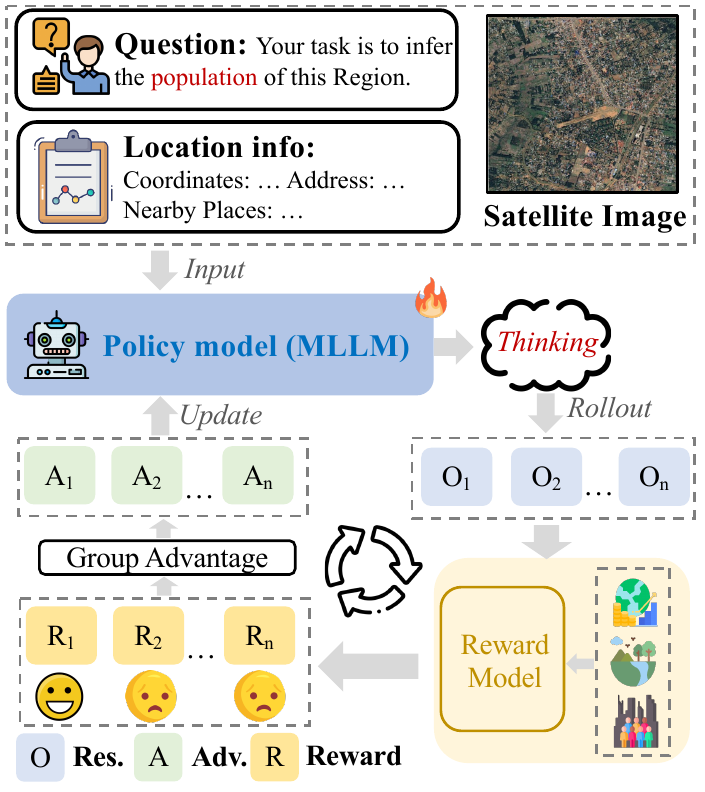}
    \caption{Training pipeline of Urban-R1.}
    \vspace{-1em}
    \label{fig:framework}
    \vspace{-0.5em}
\end{figure}

Figure~\ref{fig:framework} presents the training pipeline of \textbf{Urban-R1}, where we adopt an RL post-training framework that fine-tunes a multimodal policy model on an urban region profiling (URP) proxy task using GRPO. Concretely, for each prompt, the model generates multiple candidate answers, receives a combined reward that evaluates indicator accuracy and output parsability, and is updated with a group-relative advantage computed over the candidate set under a KL-divergence constraint to the reference policy. Aligning the policy to these task-level rewards drives evidence-grounded, geospatial-aware behavior while limiting deviation from the pretrained distribution. We then present the rationale for the URP proxy task formulation and describe the GRPO objective and the training procedure. During inference, the fine-tuned policy model directly produces structured and interpretable textual descriptions of urban indicators from multimodal inputs.

\subsection{Urban Region Profiling as a Proxy Task}
Urban region profiling aims to estimate key socioeconomic and environmental indicators of an urban region by integrating multimodal inputs. For a region $ g $, the model takes as input a satellite image $ I_g \in \mathbb{R}^{H \times W \times 3}$, location information $ L_g = (coord_g, addr_g, NP_g)$, and auxiliary text  $T_g $. Here, $coord_g$ denotes the geographic coordinates, $addr_g$ is a textual address description, and $NP_g$ represents a set of nearby named places. The model then predicts a single urban indicator $\hat{Y}_g \in \mathbb{R}$. The learning objective is to estimate the true indicators $ Y_g $ through a parametric function $ \pi $, formalized as:
\begin{equation}
\hat{Y}_g = \pi(I_g, L_g, T_g; \theta),
\end{equation}
where $\theta$ denotes the model parameters. 

We adopt URP as a proxy task because it trains models to fuse multimodal geographic evidence into calibrated estimates of key socioeconomic and environmental indicators (e.g., GDP, population, carbon emissions), which underlie many downstream applications from site selection to geo-localization. URP supplies objective, per-indicator rewards and enforceable output structure, making it practical for RL post-training, while being data-efficient: a few thousand diverse samples expose salient patterns of urban variation without costly large-scale annotation. Crucially, URP also encourages evidence-grounded outputs, yielding transferable reasoning patterns that improve calibration and help mitigate geospatial bias.

\subsection{Reinforcement Learning for URP}

\subsubsection{Reducing Geo-Bias: RL vs. SFT}
\begin{figure}[!h]
    \centering
    \includegraphics[width=1\linewidth]{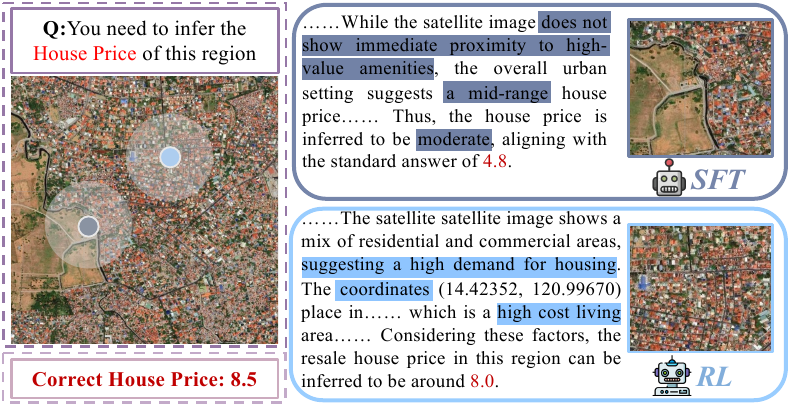}
    \caption{Comparison of house price inference by SFT and RL models, showing RL's more accurate reasoning aligned with the correct answer.}
    \vspace{-0.5em}
    \label{fig:sft_rl}
    \vspace{-0.5em}
\end{figure}

While effective in-domain, SFT suffers from critical issues leading to \textit{geo-bias}, as shown in Figure~\ref{fig:sft_rl}. It prioritizes target reproduction over verifiable geographic evidence and often generates \textit{pseudo reasoning paths}, which are plausible-sounding yet ungrounded chains of logic~\cite{chen2025sft}. For example, in house price inference, SFT focuses on superficial cues like ``lack of landmarks'' (predicting 4.8 vs. the correct 8.5), ignoring deeper geographic context~\cite{zhou2024towards}. 

RL directly addresses this mechanistic flaw by shifting the optimization objective from static token prediction to dynamic, reward-guided reasoning. Rather than merely mimicking output distributions, RL trains the model to generate predictions that are not only accurate but also grounded in verifiable spatial and geographic evidence \cite{ouyang2022training}. For instance, as shown in Figure \ref{fig:sft_rl}, when inferring house prices, the RL model does not rely on heuristic shortcuts like “lack of landmarks = low value.” Instead, it actively reasons through satellite imagery to detect mixed land use patterns and combines them with geographic coordinates to infer the local cost of living, which reflects real-world urban causality rather than statistical coincidence. By rewarding reasoning paths that align with observable features and penalizing those based on spurious correlations, RL fosters generalization across diverse urban contexts \cite{li2025recognition}.

\subsubsection{Enhancing Reasoning via GRPO}
Recent works ~\cite{liu2024deepseek, huang2025vision} have validated the effectiveness of GRPO, particularly in handling multimodal inputs and generating structured responses. Given this validation, and since GRPO’s strengths align well with URP’s need to process satellite imagery (visual) and location data (textual) for structured inferences, we adopt GRPO for RL post-training on the URP proxy task. 
\textbf{Training Pipeline.} As illustrated in Figure~\ref{fig:framework}, each training instance provides a question (e.g.,``infer the population of this region''), a satellite image, and location information. This tuple forms a URP prompt $s$. The prompt template is presented in Appendix \ref{appendix_prompt}. The Policy model receives $s$ and generate a group of $G$ candidate structured response $O(s) = \{o_{i}\}^{G}_{i=1}$.

\noindent \textbf{Reward Model.} Each candidate $o_{i}$ receives a structure reward:
\begin{equation}
    R_{i} = (1-\lambda)R_{acc}(o_{i},Y) + \lambda R_{fmt}(o_{i})
\end{equation}
where $\lambda$ is a weighting hyperparameter that balances accuracy and format fidelity. The accuracy reward $R_{acc}$ is calculated as the normalized absolute error, which is derived by taking the absolute difference between the model's inferred value from response $o_i$ and the ground-truth value $Y$, then dividing this difference by a scaling constant. The format reward $R_{fmt}$ is a binary indicator that assigns a value of 1 if $o_i$ strictly adheres to the specified answer format.

\noindent \textbf{Group Advantage}. For each prompt, which corresponds to a specific urban region, we generate multiple reasoning rollouts and compute a relative advantage for each within its geographic group. Specifically, the advantage of rollout $i$ at step $t$ is defined as:
\begin{equation}
    \hat{A}_{i,t} = \frac{R_{i}-mean(R)}{std(R)}
\end{equation}
where $R_i$ is the reward of the $i$-th rollout and $R$ denotes the set of rewards from all rollouts for the same region. This intra-group normalization enables the policy to prioritize reasoning trajectories that better capture \textit{invariant, evidence-backed spatial relationships}. By optimizing relative to peers in the same geographic context, GRPO promotes robust, generalizable reasoning and reduces reliance on spurious correlations.

\noindent \textbf{Policy update.} We optimize the KL-regularized GRPO objective:
\begin{equation}
\begin{split}
&J_{\text{GRPO}}(\theta)
= \mathbb{E}_{\,s\sim P(S),\, o_i\sim \pi_{\theta_{\text{old}}}(\cdot\mid s)}
      \frac{1}{|o_i|}\sum_{t=1}^{|o_i|}  \\
&  \!\Big[ \min \!\big(\sigma_{i,t}\,\hat A_{i,t},\;
          \operatorname{clip}(\sigma_{i,t},\,1-\epsilon,\,1+\epsilon )\,\hat A_{i,t} \!\big)
        \\
& - \beta\,D_{\mathrm{KL}}\!\big(\pi_\theta(\cdot\mid s)\,\Vert\,\pi_{\mathrm{ref}}(\cdot\mid s)\big)\Big],
\end{split}
\end{equation}
where $\displaystyle
\sigma_{i,t}=\frac{\pi_{\theta}\!\big(o_{i}(t)\mid s,\,o_{i}(<t)\big)}
{\pi_{\theta_{\text{old}}}\!\big(o_{i}(t)\mid s,\,o_{i}(<t)\big)}$
is the per-token importance ratio. The $\min$–clip term follows the
PPO-style trust-region strategy~\cite{schulman2017proximal}, which stabilizes training by preventing excessively large policy updates when
$\sigma_{i,t}$ deviates from $1$. The KL regularization term $D_{\mathrm{KL}}\!\big(\pi_\theta(\cdot\mid s)\,\Vert\,\pi_{\mathrm{ref}}(\cdot\mid s)\big)$, weighted by $\beta > 0$, encourages the updated policy $\pi_{\theta}$ to stay close to a reference model $\pi_{ref}$, typically the pretrained backbone. This preserves general capabilities while allowing controlled, reward-guided adaptation.

\begin{table}[!b]
\centering
\small
\setlength{\tabcolsep}{2.5pt}
\begin{tabular}{cccc} 
\toprule
\textbf{Features} & \makecell[c]{Contrastive\\VLMs} & \makecell[c]{SFT\\MLLMs} & Urban-R1  \\ 
\midrule
Zero-shot Inference  & \xmark & \cmark & \cmark\\
Superior Performance &\cmark &\cmark &\cmark \\
Explainablity        &\xmark   &\cmark  &\cmark  \\
Generalizability     & \xmark & \xmark &\cmark  \\
\bottomrule
\end{tabular}
\caption{Feature comparison among Contrastive Models, SFT MMLMs, and Urban-R1. \ding{51} indicates supported features; \ding{55} indicates unsupported features.}
\label{tab:feature_comparison}
\end{table}

\subsection{Comparison to Existing Arts}
As summarized in Table~\ref{tab:feature_comparison}, Urban-R1 distinguishes itself by comprehensively supporting all four critical capabilities: zero-shot inference, superior performance, explainability, and generalizability.  While Contrastive VLMs achieve strong performance but lack zero-shot capability, explainability, and generalizability, SFT MLLMs support zero-shot inference and explainability, yet still fall short in generalizability. In contrast, Urban-R1 matches or exceeds their performance while integrating all four capabilities, making it uniquely suited for real-world urban scene understanding.

\section{Experiments}

\begin{table*}[ht]
\centering
\small
\renewcommand{\arraystretch}{1.1}
\begin{tabular}{c|cc|cc|cc|cc|cc}
\toprule
\multirow{2}{*}{\large\textbf{Methods}} 
 & \multicolumn{2}{c|}{\textbf{GDP}} & \multicolumn{2}{c|}{\textbf{Carbon}} & \multicolumn{2}{c|}{\textbf{Population}} & \multicolumn{2}{c|}{\textbf{Poverty}} & \multicolumn{2}{c}{\textbf{House Price}} \\
\cmidrule(lr){2-3} \cmidrule(lr){4-5} \cmidrule(lr){6-7} \cmidrule(lr){8-9} \cmidrule(lr){10-11}
 & $\rho$ & $R^2$ & $\rho$ & $R^2$ & $\rho$ & $R^2$ & $\rho$ & $R^2$ & $\rho$ & $R^2$ \\
\midrule
\multicolumn{11}{c}{\textit{\textbf{Open-source}}} \\
\midrule
InternVL2.5-8B       & 0.717 & 0.411 & 0.610 & 0.245 & 0.758 & 0.123 & 0.485 & 0.279 & 0.003 & -1.252 \\
InternVL2.5-26B      & 0.648 & 0.310 & 0.631 & 0.250 & 0.741 & -0.160 & 0.526 & 0.237 & -0.069 & -1.292 \\
Qwen2.5-VL-7B         & 0.682 & 0.331 & 0.489 & 0.100 & 0.763 & 0.415 & 0.309 & -0.066 & -0.209 & -0.820 \\
Qwen2.5-VL-32B        & 0.718 & 0.579 & 0.724 & 0.321 & 0.832 & 0.520 & 0.694 & 0.449 & -0.063 & -1.252 \\
\midrule
\multicolumn{11}{c}{\textit{\textbf{Close-source}}} \\
\midrule
GPT-4o                & 0.765 & 0.628 & 0.749 & \underline{0.423} & \underline{0.836} & \underline{0.629} & 0.701 & 0.480 & -0.046 & -1.044 \\
Gemini-2.5-Flash      & 0.778 & 0.573 & 0.761 & -0.143 & 0.832 & 0.170 & 0.659 & 0.330 & -0.035 & -1.571 \\
\midrule
\multicolumn{11}{c}{\textit{\textbf{SFT}}} \\
\midrule
InternVL2.5-4B (SFT)& \underline{0.800} & \underline{0.701}& \underline{0.812} & \underline{0.628}  &\underline{0.824} & \underline{0.654}  & \underline{0.823} & \underline{0.640} & \underline{0.553} & 0.113\\
Qwen2.5-VL-3B (SFT)   & 0.774 & 0.668 & 0.747 & 0.507 & 0.777 & 0.581 & 0.782 & 0.558 & 0.386 & -0.121 \\
Qwen2.5-VL-7B (SFT)   & 0.785 & 0.654 & 0.714 & 0.414 & 0.805 & 0.610 & 0.772 & 0.529 & 0.527 & \underline{0.125} \\
\midrule
\multicolumn{11}{c}{\textit{\textbf{RL-Tuned}}} \\
\midrule
\rowcolor{gray!30}Urban-R1 & \textbf{0.897} & \textbf{0.836} & \textbf{0.880} & \textbf{0.785} & \textbf{0.886} & \textbf{0.775} & \textbf{0.911} & \textbf{0.777} & \textbf{0.832} & \textbf{0.723}\\
\bottomrule
\end{tabular}
\vspace{-0.5em}
\caption{Urban Region Profiling on Seen Regions. Spearman correlation $\rho$ and R$^2$ (higher $\uparrow$) for five indicators are reported. The best results are in bold, and the second-best results are underlined.}
\label{tab:main_results}
\vspace{-1em}
\end{table*}
In this section, we evaluate our proposed Urban-R1 to address the following research questions:
\begin{itemize}[leftmargin=*,topsep=0em,itemsep=0em]
    \item \textbf{RQ1}: Can Urban-R1 mitigate geospatial bias on the Urban Region Profiling task?
    \item \textbf{RQ2}: Can Urban-R1 achieve good cross-task generalization on urban downstream tasks?
    \item \textbf{RQ3}: How does each model component affect the performance of urban reasoning?
    \item \textbf{RQ4}: How is the interpretability of Urban-R1?
\end{itemize}

\subsection{Experimental Settings}
\begin{figure}[!h]
    \centering
    \includegraphics[width=1\linewidth]{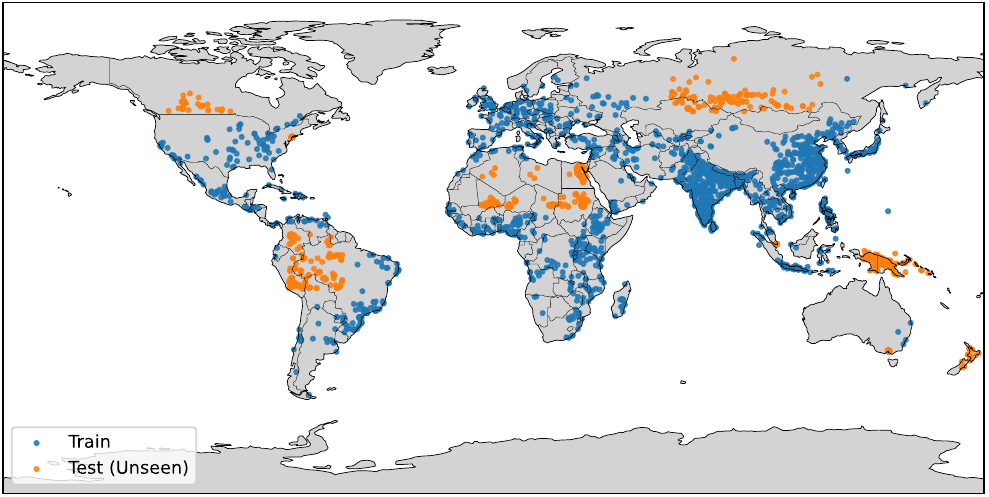}
    \caption{Geographic distribution of training (blue) and unseen test (orange) regions.}
    \label{fig:dataset_split}
\end{figure}

\textbf{Datasets.} We follow GeoLLM~\cite{manvi2023geollm} in evaluating Urban-R1 on five urban indicators: \emph{Population}, \emph{Carbon}, \emph{GDP}, \emph{Poverty}, and \emph{House Price}. To assess geo-bias mitigation, we partition the data into seen and unseen sets at the region level, as shown in Figure~\ref{fig:dataset_split}. Beyond the Urban Region Profiling task, we construct five downstream tasks to evaluate transfer and geo-bias mitigation: (1) Site Selection, (2) Scene Function, (3) Geo-localization, (4) Urban Perception, and (5) Land Use. More details of our datasets can be found in Appendix \ref{appendix_dataset}.

\noindent \textbf{Implementation Details.}
We initialize the policy with Qwen2.5-VL-7B-Instruct~\cite{bai2025qwen2}, leveraging its pre-trained multimodal instruction tuning as a robust foundation for GRPO optimization. All experiments are executed on a cluster of 4×A800 GPUs to ensure computational efficiency. To facilitate better adaptation of the visual backbone to satellite imagery, the vision encoder remains trainable throughout the RL process. For GRPO training, we adopt the EasyR1
framework with the following RL hyperparameters: a global batch size 128, a rollout batch size 256, and a rollout temperature 1.0, and a learning rate set to $1 \times 10^{-6}$.

\noindent \textbf{Baselines.}
We compare Urban-R1 with the following three model families:
\begin{itemize}[leftmargin=*,topsep=0em,itemsep=0em]
    \item \textbf{Open-source MLLMs (in zero-shot settings):} InternVL2.5-8B~\cite{chen2024expanding}, InternVL2.5-26B~\cite{chen2024expanding}, Qwen2.5-VL-7B~\cite{bai2025qwen2}, Qwen2.5-VL-32B~\cite{bai2025qwen2}. These models are evaluated without task-specific tuning, using a unified input prompt.
    \item \textbf{Closed-source MLLMs (zero-shot):} GPT-4o~\cite{hurst2024gpt} and Gemini-2.5-Flash~\cite{comanici2025gemini}. We submit the identical prompts and enforce the same output schema.
    \item \textbf{SFT baselines:} We conduct full SFT on base models including InternVL2.5-4B, Qwen2.5-VL-3B, and Qwen2.5-VL-7B using the URP training split, with identical image-location inputs and target schema as Urban-R1. Full implementation details are provided in Appendix~\ref{app:implementation}.
\end{itemize}

\noindent \textbf{Evaluation Metrics.}
Following prior work \cite{hao2024urbanvlp,yan2024urbanclip}, we report the coefficient of determination $R^{2}$:
\begin{equation}
\footnotesize
    R^2 \;=\; 1 \;-\; 
\frac{\sum_{g=1}^{N}\big(\hat{y}_{g}-y_{g}\big)^2}
     {\sum_{g=1}^{N}\big(\hat{y}_{g}-\bar{y}\big)^2},
\end{equation}
where $\hat{y}_{g}$ denotes ground truth, $y_{g}$ indicates predictions, and $\bar{y}$ is the mean of $\{y_g\}^{N}_{g=1}$.
To quantify geo-bias, we use Spearman rank correlation $\rho = corr\big(rank(Y),rank(\hat{Y})\big)$ between ground truth and predicted value rankings (higher indicates better geo-consistency, i.e., lower bias).
For downstream tasks, we use classification accuracy.

\subsection{Mitigating Geo-bias (RQ1)}
To address RQ1, whether Urban-R1 can perform well on the Urban Region Profiling task and solve geospatial bias in unseen regions, we evaluate Spearman correlation $\rho$ and $R^2$ across five urban indicators in both seen and unseen regions.

\begin{figure*}[!ht]
    \centering
    \includegraphics[width=1\linewidth]{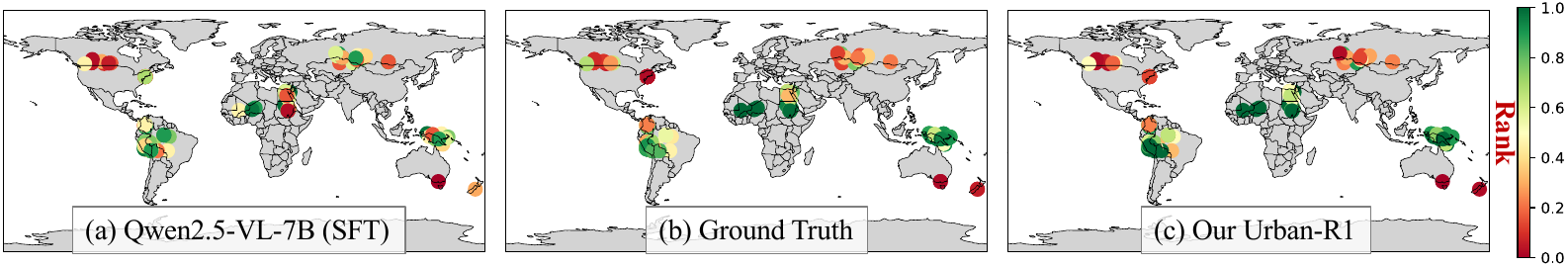}
    \vspace{-1em}
    \caption{Poverty prediction ranks on unseen regions, with  indicating higher ranks and red representing lower ranks. Notably, Urban-R1 aligns more closely with the ground truth pattern compared to the SFT baseline.}
    \label{fig:experiment_bias}
    \vspace{-1em}
\end{figure*}

\subsubsection{Superior Performance on Seen Regions}
For seen regions, Urban-R1 achieves the best performance across all indicators as shown in Table~\ref{tab:main_results} while SFT variants show only modest and inconsistent gains, and other MLLMs underperform due to generic pretraining that favors plausibility over geospatially grounded numerically calibrated reasoning and due to SFT reliance on memorization, which leads to unstable or negative House Price predictions. Urban-R1 overcomes these issues via GRPO, a relative policy optimization method that samples multiple responses per prompt, scores them using rewards that account for both accuracy and format, and reinforces the candidate that performs best within each group. 

\subsubsection{Results on Unseen Regions}
We assess geo-bias on unseen regions using Spearman's rank correlation between ground-truth and predicted rankings. Table~\ref{tab:unseen_results} shows that Urban-R1 attains the highest positive correlations across all five indicators, substantially outperforming the baselines. In particular, GPT-4o exhibits near-zero or even negative agreement on context-sensitive metrics such as House Price, with a Spearman correlation of -0.009. Meanwhile, 7B-SFT achieves modest positive performance, for example, a correlation of 0.363 for House Price, but falls short on other indicators like Poverty, where it reaches 0.808 compared to Urban-R1’s 0.915.
\begin{table}[!t]
\centering
\small
\setlength{\tabcolsep}{2.5pt}
\renewcommand{\arraystretch}{1.20}
\resizebox{\columnwidth}{!}{
\begin{tabular}{l|c|c|c|c}
\toprule
\makecell[l]{\textbf{Indicator $(\rho)$ }} & \makecell{\textbf{GPT-4o}} & \makecell{\textbf{InternVL (SFT)}} & \makecell{\textbf{Qwen2.5 (SFT)}} & \makecell{\textbf{Urban-R1}} \\
\hline
Carbon     & 0.728 & 0.448 & 0.691 & \textbf{0.839} \\
Poverty   & 0.630 & 0.489 & 0.808 & \textbf{0.915} \\
Population  & 0.899 & 0.382 & 0.840 & \textbf{0.907} \\
GDP         & 0.734 & 0.028 & 0.738 & \textbf{0.833} \\
House Price & -0.009 & 0.317 & 0.363 & \textbf{0.765} \\
\bottomrule
\end{tabular}}
\vspace{-1em}
\caption{Spearman correlation coefficients $\rho$ of different models on \textbf{unseen regions} across five indicators. InternVL: InternVL2.5-4B; Qwen2.5: Qwen2.5-VL-7B.}
\label{tab:unseen_results}
\vspace{-1em}
\end{table}

These discrepancies arise because GPT-4o, despite its scale, over-relies on broad global priors from pre-training data sourced from economically developed regions, leading to homogenized predictions that ignore local geographic differences. This data bias causes systematic underestimation of economic outputs in less-represented regions and overemphasis on developed-world patterns. Conversely, 7B-SFT's smaller capacity and supervised fine-tuning exacerbate overfitting to seen-region, yielding inconsistent generalization on long-tail unseen areas. The example rank map in Figure~\ref{fig:experiment_bias} echoes this pattern: Urban-R1 reproduces the spatial ordering, while the SFT baseline collapses toward global priors and misorders long-tail regions. These gains stem from geospatial-aware reasoning rather than merely memorizing the training data.

\begin{figure}[!b]
    \vspace{-1.3em}
    \centering
    \includegraphics[width=0.97\linewidth]{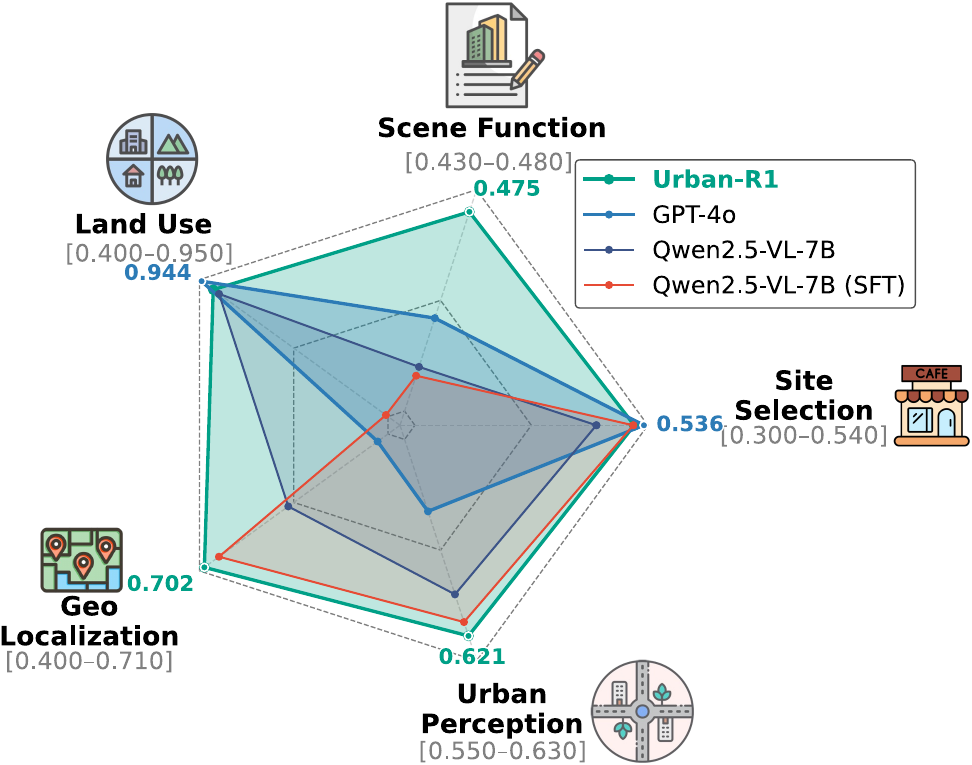}
    \vspace{-0.7em}
    \caption{Accuracy radar across five downstream tasks.}
    \label{fig:downstream}
\end{figure}
\subsection{Results on Downstream Tasks (RQ2)}
To address RQ2, we evaluate five downstream tasks, spanning diverse urban reasoning scenarios:
\begin{itemize}[leftmargin=*,topsep=0em,itemsep=-0.1em]
    \item \textbf{Site Selection}: Determine if coordinates is suitable for building specific commercial establishments (e.g., KFC, Starbucks) based on POIs.
    \item \textbf{Scene Function}: Select the satellite image (from urban-area images) that most likely contains the largest number of specified food-related POIs (e.g., restaurants, bakeries, fast-food venues).
    \item \textbf{Land Use}: Classify the most probable land-use type of a region (e.g., Residential, Grass, Retail) from a satellite image.
    \item \textbf{Geo-Localization}: Identify which coordinate (among four candidates) corresponds to the location in a satellite image.
    \item \textbf{Urban Perception}: Make perceptual judgments about urban scenes (e.g., which place looks more livable and safer?).
\end{itemize}

As shown in Figure~\ref{fig:downstream}, Urban-R1 achieves strong performance across all downstream tasks: it ranks first in Scene Function and Geo-localization, and remains highly competitive in other tasks. In contrast, the SFT variant exhibits compromised performance: for example, its accuracy in Land Use drops, even lagging behind the base Qwen2.5-VL-7B in this task. This reveals that supervised fine-tuning can lead to task-specific overfitting, undermining generalization on downstream urban tasks. Meanwhile, Urban-R1’s performance is either the best or close to that of the closed-source model (GPT-4o) across these tasks. Notably, this competitiveness is more valuable considering Urban-R1’s open-source attribute, which lowers the barrier for practical applications in urban research. Overall, these results illustrate that the RL approach yields geography-aware reasoning that boosts accuracy across diverse urban tasks. 

\subsection{Ablation Study (RQ3)}
\begin{figure}[!t]
    \centering
    \includegraphics[width=1\linewidth]{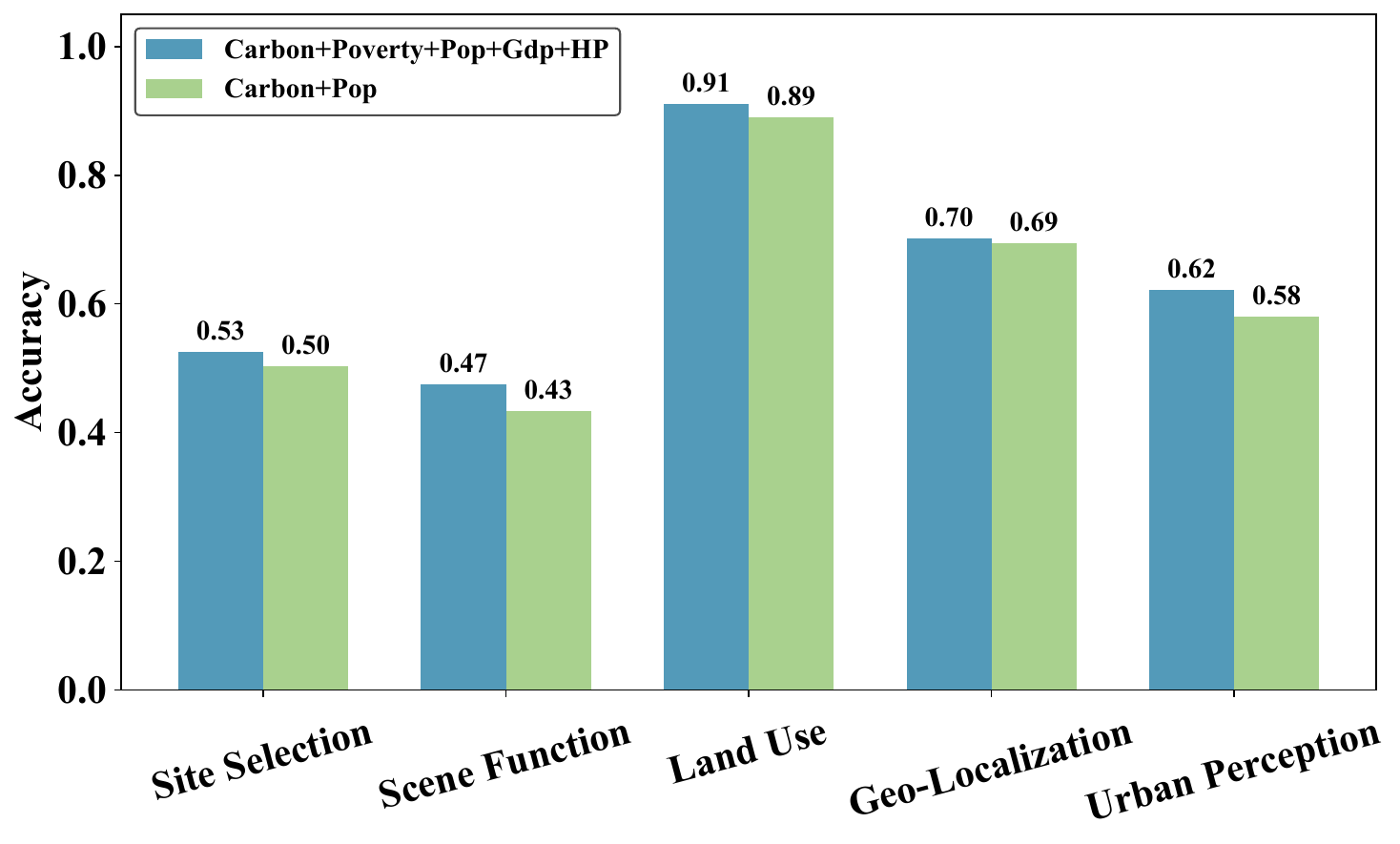}
    \vspace{-0.5em}
    \caption{Ablation study on using different urban indicator sets for training across urban downstream tasks. HP: House Price; Pop: Population.}
    \label{fig:dataset_ablation}
    \vspace{-1em}
\end{figure}
To investigate factors affecting urban intelligence reasoning, we conduct two ablation studies: one on urban indicator types’ impact on downstream performance, the other on input modalities’ role.

\textbf{Urban Indicator Types.} We examine how the inclusion of diverse urban indicators affects performance across downstream tasks. Specifically, we compare a model using a rich set of indicators (Carbon, Poverty, Population, GDP, House Price) with a variant that uses only two indicators (Carbon, Population). As shown in Figure~\ref{fig:dataset_ablation}, the model with richer indicators outperforms the simplified variant across all tasks. For example, Urban Perception data shows the comprehensive-indicator model hits 0.62 accuracy, and the two-indicator variant scores 0.58, which demonstrates diverse socioeconomic and physical indicators boost the model’s urban complexity capture and downstream performance.

\textbf{Input Modalities.} To empirically evaluate the contribution of satellite imagery and geographic text features to the reasoning mechanism, we conduct a comparative analysis between two model variants: (1) \textbf{w/o Image}: only using location coordinates and geographic textual descriptors, without satellite imagery; (2) \textbf{w/o Text}: only using satellite imagery, without location or textual geographic information. As Table~\ref{tab:ablation} illustrates, removing either modality degrades performance in seen regions. Without image input, the model fails to capture fine-grained physical features, leading to severe drops for indicators like Carbon and Poverty. Without text input, the model loses geographic contextual grounding, causing collapses for context-sensitive indicators such as House Price (from 0.723 to $-$0.691). In contrast, Urban-R1 maintains robust performance across all indicators.

\begin{table}[!t]
\centering
\small 
\setlength{\tabcolsep}{6.5pt}
\renewcommand{\arraystretch}{1.05}
\begin{tabular}{l|c|c|c}
\hline
\textbf{Indicator ($R^2$)} & \textbf{w/o Image} & \textbf{w/o Text} & \textbf{Urban-R1} \\
\cline{2-4}
\hline
GDP        & 0.338   & 0.824   &\textbf{0.836} \\
Carbon     & 0.241  & 0.712   & \textbf{0.785}  \\
Population  & 0.577   & 0.767   &\textbf{ 0.775}  \\
Poverty & 0.461 & 0.741  & \textbf{0.777}  \\
House Price & -0.311  & -0.691 & \textbf{0.723} \\
\hline
\end{tabular}
\vspace{-0.5em}
\caption{Urban-R1 vs. Urban-R1 without satellite imagery (\textbf{w/o Image}) or geographic texts (\textbf{w/o Text}).}
\label{tab:ablation}
\vspace{-1.5em}
\end{table}

\subsection{Interpretability Study (RQ4)}

\begin{figure}[!h]
    \centering
    \vspace{-0.5em}
    \includegraphics[width=1\linewidth]{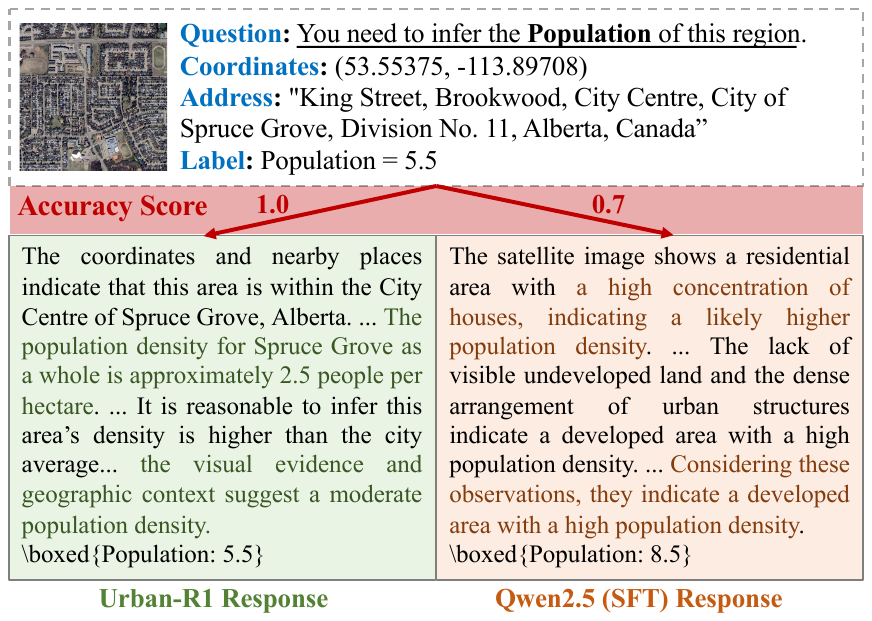}
    \vspace{-1em}
    \caption{A case study on an unseen region.}
    \label{fig:case}
    \vspace{-1em}
\end{figure}

To evaluate the interpretability of Urban-R1, we present a representative example from the unseen-region test set (Figure \ref{fig:case}). The task involves estimating the population of a Canadian region. Urban-R1 correctly infers a value of 5.5 by grounding its reasoning in quantifiable geographic evidence, such as the estimated population density of Spruce Grove ($\sim$2.5 people per hectare), and visual cues from satellite imagery. In contrast, Qwen2.5 (SFT) relies on vague qualitative descriptions and produces an erroneous estimate of 8.5. This comparison demonstrates how Urban-R1’s reinforcement-aligned reasoning yields more transparent, evidence-based interpretations of urban indicators.

\vspace{-0.2em}
\section{Conclusion and Future Work}

\vspace{-0.3em}
This paper presents Urban-R1, a reinforced MLLM for urban general intelligence.
Trained with GRPO on the urban region profiling proxy task, Urban-R1 learns geography-invariant reasoning patterns that effectively mitigate geospatial bias and sustain strong performance on unseen regions.
Across urban reasoning benchmarks, it outperforms SFT baselines and leading closed-source models, showing the promise of reinforcement learning for fair and generalizable urban intelligence.
Future work will extend Urban-R1 with tool-use and interaction capabilities, enabling dynamic invocation of urban analytics and real-time monitoring tools to address more complex decision-making scenarios. 

\clearpage

\bibliography{acl_ref}

\begin{thebibliography}{47}
\providecommand{\natexlab}[1]{#1}

\bibitem[{Bai et~al.(2025)Bai, Chen, Liu, Wang, Ge, Song, Dang, Wang, Wang, Tang et~al.}]{bai2025qwen2}
Shuai Bai, Keqin Chen, Xuejing Liu, Jialin Wang, Wenbin Ge, Sibo Song, Kai Dang, Peng Wang, Shijie Wang, Jun Tang, and 1 others. 2025.
\newblock Qwen2. 5-vl technical report.
\newblock \emph{arXiv preprint arXiv:2502.13923}.

\bibitem[{Bai et~al.(2022)Bai, Jones, Ndousse, Askell, Chen, DasSarma, Drain, Fort, Ganguli, Henighan et~al.}]{bai2022training}
Yuntao Bai, Andy Jones, Kamal Ndousse, Amanda Askell, Anna Chen, Nova DasSarma, Dawn Drain, Stanislav Fort, Deep Ganguli, Tom Henighan, and 1 others. 2022.
\newblock Training a helpful and harmless assistant with reinforcement learning from human feedback.
\newblock \emph{arXiv preprint arXiv:2204.05862}.

\bibitem[{Bjorck et~al.(2021)Bjorck, Rappazzo, Shi, Brown-Lima, Dean, Fuller, and Gomes}]{bjorck2021accelerating}
Johan Bjorck, Brendan~H Rappazzo, Qinru Shi, Carrie Brown-Lima, Jennifer Dean, Angela Fuller, and Carla Gomes. 2021.
\newblock Accelerating ecological sciences from above: Spatial contrastive learning for remote sensing.
\newblock In \emph{Proceedings of the AAAI Conference on Artificial Intelligence}, volume~35, pages 14711--14720.

\bibitem[{Cao et~al.(2024)Cao, Zhao, Cheng, Shu, Chen, Liu, Liang, Zhao, Yan, and Li}]{cao2024survey}
Yuji Cao, Huan Zhao, Yuheng Cheng, Ting Shu, Yue Chen, Guolong Liu, Gaoqi Liang, Junhua Zhao, Jinyue Yan, and Yun Li. 2024.
\newblock Survey on large language model-enhanced reinforcement learning: Concept, taxonomy, and methods.
\newblock \emph{IEEE Transactions on Neural Networks and Learning Systems}.

\bibitem[{Chen et~al.(2025)Chen, Tu, Wang, Liu, Tang, Du, Zhou, and Xie}]{chen2025sft}
Hardy Chen, Haoqin Tu, Fali Wang, Hui Liu, Xianfeng Tang, Xinya Du, Yuyin Zhou, and Cihang Xie. 2025.
\newblock Sft or rl? an early investigation into training r1-like reasoning large vision-language models.
\newblock \emph{arXiv preprint arXiv:2504.11468}.

\bibitem[{Chen et~al.(2024)Chen, Wang, Cao, Liu, Gao, Cui, Zhu, Ye, Tian, Liu et~al.}]{chen2024expanding}
Zhe Chen, Weiyun Wang, Yue Cao, Yangzhou Liu, Zhangwei Gao, Erfei Cui, Jinguo Zhu, Shenglong Ye, Hao Tian, Zhaoyang Liu, and 1 others. 2024.
\newblock Expanding performance boundaries of open-source multimodal models with model, data, and test-time scaling.
\newblock \emph{arXiv preprint arXiv:2412.05271}.

\bibitem[{Chu et~al.(2025)Chu, Zhai, Yang, Tong, Xie, Schuurmans, Le, Levine, and Ma}]{chu2025sft}
Tianzhe Chu, Yuexiang Zhai, Jihan Yang, Shengbang Tong, Saining Xie, Dale Schuurmans, Quoc~V Le, Sergey Levine, and Yi~Ma. 2025.
\newblock Sft memorizes, rl generalizes: A comparative study of foundation model post-training.
\newblock \emph{arXiv preprint arXiv:2501.17161}.

\bibitem[{Comanici et~al.(2025)Comanici, Bieber, Schaekermann, Pasupat, Sachdeva, Dhillon, Blistein, Ram, Zhang, Rosen et~al.}]{comanici2025gemini}
Gheorghe Comanici, Eric Bieber, Mike Schaekermann, Ice Pasupat, Noveen Sachdeva, Inderjit Dhillon, Marcel Blistein, Ori Ram, Dan Zhang, Evan Rosen, and 1 others. 2025.
\newblock Gemini 2.5: Pushing the frontier with advanced reasoning, multimodality, long context, and next generation agentic capabilities.
\newblock \emph{arXiv preprint arXiv:2507.06261}.

\bibitem[{Dubey et~al.(2016)Dubey, Naik, Parikh, Raskar, and Hidalgo}]{dubey2016deep}
Abhimanyu Dubey, Nikhil Naik, Devi Parikh, Ramesh Raskar, and C{\'e}sar~A Hidalgo. 2016.
\newblock Deep learning the city: Quantifying urban perception at a global scale.
\newblock In \emph{European conference on computer vision}, pages 196--212. Springer.

\bibitem[{Feng et~al.(2025{\natexlab{a}})Feng, Liu, Du, Guo, Lin, and Li}]{feng2025citygpt}
Jie Feng, Tianhui Liu, Yuwei Du, Siqi Guo, Yuming Lin, and Yong Li. 2025{\natexlab{a}}.
\newblock Citygpt: Empowering urban spatial cognition of large language models.
\newblock In \emph{Proceedings of the 31st ACM SIGKDD Conference on Knowledge Discovery and Data Mining V. 2}, pages 591--602.

\bibitem[{Feng et~al.(2025{\natexlab{b}})Feng, Wang, Liu, Xi, and Li}]{feng2025urbanllava}
Jie Feng, Shengyuan Wang, Tianhui Liu, Yanxin Xi, and Yong Li. 2025{\natexlab{b}}.
\newblock Urbanllava: A multi-modal large language model for urban intelligence with spatial reasoning and understanding.
\newblock \emph{arXiv preprint arXiv:2506.23219}.

\bibitem[{Gao et~al.(2023)Gao, Schulman, and Hilton}]{gao2023scaling}
Leo Gao, John Schulman, and Jacob Hilton. 2023.
\newblock Scaling laws for reward model overoptimization.
\newblock In \emph{International Conference on Machine Learning}, pages 10835--10866. PMLR.

\bibitem[{Hao et~al.(2024)Hao, Chen, Yan, Zhong, Wang, Wen, and Liang}]{hao2024urbanvlp}
Xixuan Hao, Wei Chen, Yibo Yan, Siru Zhong, Kun Wang, Qingsong Wen, and Yuxuan Liang. 2024.
\newblock Urbanvlp: A multi-granularity vision-language pre-trained foundation model for urban indicator prediction.
\newblock \emph{arXiv e-prints}, pages arXiv--2403.

\bibitem[{He et~al.(2018)He, Yang, Zhang, and Zhang}]{he2018perceiving}
Zhiyuan He, Su~Yang, Weishan Zhang, and Jiulong Zhang. 2018.
\newblock Perceiving commerial activeness over satellite images.
\newblock In \emph{Companion Proceedings of the The Web Conference 2018}, pages 387--394.

\bibitem[{Huang et~al.(2021)Huang, Wang, Sheng, Ng, and Rajagopal}]{huang2021m3g}
Tianyuan Huang, Zhecheng Wang, Hao Sheng, Andrew~Y Ng, and Ram Rajagopal. 2021.
\newblock M3g: Learning urban neighborhood representation from multi-modal multi-graph.
\newblock In \emph{Proceedings of the DeepSpatial 2021: 2nd ACM KDD Workshop on Deep Learning for Spatio-Temporal Data, Applications and Systems}.

\bibitem[{Huang et~al.(2025)Huang, Jia, Zhai, Cao, Ye, Zhao, Xu, Hu, and Lin}]{huang2025vision}
Wenxuan Huang, Bohan Jia, Zijie Zhai, Shaosheng Cao, Zheyu Ye, Fei Zhao, Zhe Xu, Yao Hu, and Shaohui Lin. 2025.
\newblock Vision-r1: Incentivizing reasoning capability in multimodal large language models.
\newblock \emph{arXiv preprint arXiv:2503.06749}.

\bibitem[{Hurst et~al.(2024)Hurst, Lerer, Goucher, Perelman, Ramesh, Clark, Ostrow, Welihinda, Hayes, Radford et~al.}]{hurst2024gpt}
Aaron Hurst, Adam Lerer, Adam~P Goucher, Adam Perelman, Aditya Ramesh, Aidan Clark, AJ~Ostrow, Akila Welihinda, Alan Hayes, Alec Radford, and 1 others. 2024.
\newblock Gpt-4o system card.
\newblock \emph{arXiv preprint arXiv:2410.21276}.

\bibitem[{Jean et~al.(2019)Jean, Wang, Samar, Azzari, Lobell, and Ermon}]{jean2019tile2vec}
Neal Jean, Sherrie Wang, Anshul Samar, George Azzari, David Lobell, and Stefano Ermon. 2019.
\newblock Tile2vec: Unsupervised representation learning for spatially distributed data.
\newblock In \emph{Proceedings of the AAAI Conference on Artificial Intelligence}, volume~33, pages 3967--3974.

\bibitem[{Kang et~al.(2020)Kang, Fernandez-Beltran, Duan, Liu, and Plaza}]{kang2020deep}
Jian Kang, Ruben Fernandez-Beltran, Puhong Duan, Sicong Liu, and Antonio~J Plaza. 2020.
\newblock Deep unsupervised embedding for remotely sensed images based on spatially augmented momentum contrast.
\newblock \emph{IEEE Transactions on Geoscience and Remote Sensing}, 59(3):2598--2610.

\bibitem[{Kuckreja et~al.(2024)Kuckreja, Danish, Naseer, Das, Khan, and Khan}]{kuckreja2024geochat}
Kartik Kuckreja, Muhammad~Sohail Danish, Muzammal Naseer, Abhijit Das, Salman Khan, and Fahad~Shahbaz Khan. 2024.
\newblock Geochat: Grounded large vision-language model for remote sensing.
\newblock In \emph{Proceedings of the IEEE/CVF Conference on Computer Vision and Pattern Recognition}, pages 27831--27840.

\bibitem[{Lai et~al.(2025)Lai, Xu, Zhang, Liu, and Xiong}]{lai2025llmlight}
Siqi Lai, Zhao Xu, Weijia Zhang, Hao Liu, and Hui Xiong. 2025.
\newblock Llmlight: Large language models as traffic signal control agents.
\newblock In \emph{Proceedings of the 31st ACM SIGKDD Conference on Knowledge Discovery and Data Mining V. 1}, pages 2335--2346.

\bibitem[{Law et~al.(2019)Law, Paige, and Russell}]{law2019take}
Stephen Law, Brooks Paige, and Chris Russell. 2019.
\newblock Take a look around: using street view and satellite images to estimate house prices.
\newblock \emph{ACM Transactions on Intelligent Systems and Technology (TIST)}, 10(5):1--19.

\bibitem[{Li et~al.(2025)Li, Zhou, Liang, Tsung, and Wei}]{li2025recognition}
Ling Li, Yao Zhou, Yuxuan Liang, Fugee Tsung, and Jiaheng Wei. 2025.
\newblock Recognition through reasoning: Reinforcing image geo-localization with large vision-language models.
\newblock \emph{arXiv preprint arXiv:2506.14674}.

\bibitem[{Li et~al.(2024)Li, Xia, Tang, Xu, Shi, Xia, Yin, and Huang}]{li2024urbangpt}
Zhonghang Li, Lianghao Xia, Jiabin Tang, Yong Xu, Lei Shi, Long Xia, Dawei Yin, and Chao Huang. 2024.
\newblock Urbangpt: Spatio-temporal large language models.
\newblock In \emph{Proceedings of the 30th ACM SIGKDD Conference on Knowledge Discovery and Data Mining}, pages 5351--5362.

\bibitem[{Liang et~al.(2025)Liang, Wen, Xia, Jin, Yang, Salim, Wen, Pan, and Cong}]{liang2025foundation}
Yuxuan Liang, Haomin Wen, Yutong Xia, Ming Jin, Bin Yang, Flora Salim, Qingsong Wen, Shirui Pan, and Gao Cong. 2025.
\newblock Foundation models for spatio-temporal data science: A tutorial and survey.
\newblock In \emph{Proceedings of the 31st ACM SIGKDD Conference on Knowledge Discovery and Data Mining V. 2}, pages 6063--6073.

\bibitem[{Liu et~al.(2024)Liu, Feng, Xue, Wang, Wu, Lu, Zhao, Deng, Zhang, Ruan et~al.}]{liu2024deepseek}
Aixin Liu, Bei Feng, Bing Xue, Bingxuan Wang, Bochao Wu, Chengda Lu, Chenggang Zhao, Chengqi Deng, Chenyu Zhang, Chong Ruan, and 1 others. 2024.
\newblock Deepseek-v3 technical report.
\newblock \emph{arXiv preprint arXiv:2412.19437}.

\bibitem[{Liu et~al.(2025)Liu, Yang, Qian, Yin, Wang, Li, Liu, Zhai, Liu, and Zhang}]{liu2025reinforcement}
Keliang Liu, Dingkang Yang, Ziyun Qian, Weijie Yin, Yuchi Wang, Hongsheng Li, Jun Liu, Peng Zhai, Yang Liu, and Lihua Zhang. 2025.
\newblock Reinforcement learning meets large language models: A survey of advancements and applications across the llm lifecycle.
\newblock \emph{arXiv preprint arXiv:2509.16679}.

\bibitem[{{LLaMA Factory Team}(2024)}]{llamafactory}
{LLaMA Factory Team}. 2024.
\newblock {LLaMA Factory: Easy-to-use LLM Training Framework}.
\newblock \url{https://llamafactory.readthedocs.io/zh-cn/latest/}.

\bibitem[{Manvi et~al.(2024)Manvi, Khanna, Burke, Lobell, and Ermon}]{manvi2024large}
Rohin Manvi, Samar Khanna, Marshall Burke, David Lobell, and Stefano Ermon. 2024.
\newblock Large language models are geographically biased.
\newblock \emph{arXiv preprint arXiv:2402.02680}.

\bibitem[{Manvi et~al.(2023)Manvi, Khanna, Mai, Burke, Lobell, and Ermon}]{manvi2023geollm}
Rohin Manvi, Samar Khanna, Gengchen Mai, Marshall Burke, David Lobell, and Stefano Ermon. 2023.
\newblock Geollm: Extracting geospatial knowledge from large language models.
\newblock \emph{arXiv preprint arXiv:2310.06213}.

\bibitem[{{OpenGVLab}(2024)}]{internvl}
{OpenGVLab}. 2024.
\newblock {InternVL: Closing the Gap to Commercial Multimodal Models}.
\newblock \url{https://internvl.readthedocs.io/en/latest/}.

\bibitem[{Ouyang et~al.(2022)Ouyang, Wu, Jiang, Almeida, Wainwright, Mishkin, Zhang, Agarwal, Slama, Ray et~al.}]{ouyang2022training}
Long Ouyang, Jeffrey Wu, Xu~Jiang, Diogo Almeida, Carroll Wainwright, Pamela Mishkin, Chong Zhang, Sandhini Agarwal, Katarina Slama, Alex Ray, and 1 others. 2022.
\newblock Training language models to follow instructions with human feedback.
\newblock \emph{Advances in neural information processing systems}, 35:27730--27744.

\bibitem[{Park et~al.(2022)Park, Han, Ahn, Kim, Yang, Lee, Hong, Kim, Park, Yang et~al.}]{park2022learning}
Sungwon Park, Sungwon Han, Donghyun Ahn, Jaeyeon Kim, Jeasurk Yang, Susang Lee, Seunghoon Hong, Jihee Kim, Sangyoon Park, Hyunjoo Yang, and 1 others. 2022.
\newblock Learning economic indicators by aggregating multi-level geospatial information.
\newblock In \emph{Proceedings of the AAAI Conference on Artificial Intelligence}, volume~36, pages 12053--12061.

\bibitem[{Pino(2018)}]{melbourne_housing}
Anthony Pino. 2018.
\newblock {Melbourne Housing Market Dataset}.
\newblock Kaggle Dataset. \url{https://www.kaggle.com/datasets/anthonypino/melbourne-housing-market}.

\bibitem[{Ramamurthy et~al.(2022)Ramamurthy, Ammanabrolu, Brantley, Hessel, Sifa, Bauckhage, Hajishirzi, and Choi}]{ramamurthy2022reinforcement}
Rajkumar Ramamurthy, Prithviraj Ammanabrolu, Kiant{\'e} Brantley, Jack Hessel, Rafet Sifa, Christian Bauckhage, Hannaneh Hajishirzi, and Yejin Choi. 2022.
\newblock Is reinforcement learning (not) for natural language processing: Benchmarks, baselines, and building blocks for natural language policy optimization.
\newblock \emph{arXiv preprint arXiv:2210.01241}.

\bibitem[{Schulman et~al.(2017)Schulman, Wolski, Dhariwal, Radford, and Klimov}]{schulman2017proximal}
John Schulman, Filip Wolski, Prafulla Dhariwal, Alec Radford, and Oleg Klimov. 2017.
\newblock Proximal policy optimization algorithms.
\newblock \emph{arXiv preprint arXiv:1707.06347}.

\bibitem[{Wang et~al.(2020)Wang, Li, and Rajagopal}]{wang2020urban2vec}
Zhecheng Wang, Haoyuan Li, and Ram Rajagopal. 2020.
\newblock Urban2vec: Incorporating street view imagery and pois for multi-modal urban neighborhood embedding.
\newblock In \emph{Proceedings of the AAAI Conference on Artificial Intelligence}, volume~34, pages 1013--1020.

\bibitem[{Withana(2023)}]{ny_housing}
Nelgiri Withana. 2023.
\newblock {New York Housing Market Dataset}.
\newblock Kaggle Dataset. \url{https://www.kaggle.com/datasets/nelgiriyewithana/new-york-housing-market}.

\bibitem[{Xi(2022)}]{singapore_housing}
Lize Xi. 2022.
\newblock {Singapore Public Housing Dataset}.
\newblock Kaggle Dataset. \url{https://www.kaggle.com/datasets/lizexi/singapore-public-housing-dataset}.

\bibitem[{Xi et~al.(2022)Xi, Li, Wang, Li, Tarkoma, and Hui}]{xi2022beyond}
Yanxin Xi, Tong Li, Huandong Wang, Yong Li, Sasu Tarkoma, and Pan Hui. 2022.
\newblock Beyond the first law of geography: Learning representations of satellite imagery by leveraging point-of-interests.
\newblock In \emph{Proceedings of the ACM Web Conference 2022}, pages 3308--3316.

\bibitem[{Xia et~al.(2017)Xia, Hu, Hu, Shi, Bai, Zhong, Zhang, and Lu}]{xia2017aid}
Gui-Song Xia, Jingwen Hu, Fan Hu, Baoguang Shi, Xiang Bai, Yanfei Zhong, Liangpei Zhang, and Xiaoqiang Lu. 2017.
\newblock Aid: A benchmark data set for performance evaluation of aerial scene classification.
\newblock \emph{IEEE Transactions on Geoscience and Remote Sensing}, 55(7):3965--3981.

\bibitem[{Xiao et~al.(2024)Xiao, Zhou, Xiao, Huang, and Xiong}]{xiao2024refound}
Congxi Xiao, Jingbo Zhou, Yixiong Xiao, Jizhou Huang, and Hui Xiong. 2024.
\newblock Refound: Crafting a foundation model for urban region understanding upon language and visual foundations.
\newblock In \emph{Proceedings of the 30th ACM SIGKDD Conference on Knowledge Discovery and Data Mining}, pages 3527--3538.

\bibitem[{Yan et~al.(2024)Yan, Wen, Zhong, Chen, Chen, Wen, Zimmermann, and Liang}]{yan2024urbanclip}
Yibo Yan, Haomin Wen, Siru Zhong, Wei Chen, Haodong Chen, Qingsong Wen, Roger Zimmermann, and Yuxuan Liang. 2024.
\newblock Urbanclip: Learning text-enhanced urban region profiling with contrastive language-image pretraining from the web.
\newblock In \emph{Proceedings of the ACM Web Conference 2024}, pages 4006--4017.

\bibitem[{Yeh et~al.(2020)Yeh, Perez, Driscoll, Azzari, Tang, Lobell, Ermon, and Burke}]{yeh2020using}
Christopher Yeh, Anthony Perez, Anne Driscoll, George Azzari, Zhongyi Tang, David Lobell, Stefano Ermon, and Marshall Burke. 2020.
\newblock Using publicly available satellite imagery and deep learning to understand economic well-being in africa.
\newblock \emph{Nature communications}, 11(1):2583.

\bibitem[{Zhang et~al.(2024)Zhang, Han, Xu, Ni, Liu, and Xiong}]{zhang2024towards}
Weijia Zhang, Jindong Han, Zhao Xu, Hang Ni, Hao Liu, and Hui Xiong. 2024.
\newblock Towards urban general intelligence: A review and outlook of urban foundation models.
\newblock \emph{arXiv preprint arXiv:2402.01749}.

\bibitem[{Zhou et~al.(2024)Zhou, Feng, Ke, Jiang, Yan, Yang, and Zhang}]{zhou2024towards}
Yue Zhou, Litong Feng, Yiping Ke, Xue Jiang, Junchi Yan, Xue Yang, and Wayne Zhang. 2024.
\newblock Towards vision-language geo-foundation model: A survey.
\newblock \emph{arXiv preprint arXiv:2406.09385}.

\bibitem[{Zou et~al.(2025)Zou, Yang, Chen, Hao, Chen, Huang, and Liang}]{zou2025traffic}
Xingchen Zou, Yuhao Yang, Zheng Chen, Xixuan Hao, Yiqi Chen, Chao Huang, and Yuxuan Liang. 2025.
\newblock Traffic-r1: Reinforced llms bring human-like reasoning to traffic signal control systems.
\newblock \emph{arXiv preprint arXiv:2508.02344}.

\end{thebibliography}

\clearpage
\appendix 
\section{Appendix}
\subsection{Prompt Template}\label{appendix_prompt}
\begin{figure}[!h]
    \centering
    \includegraphics[width=1\linewidth]{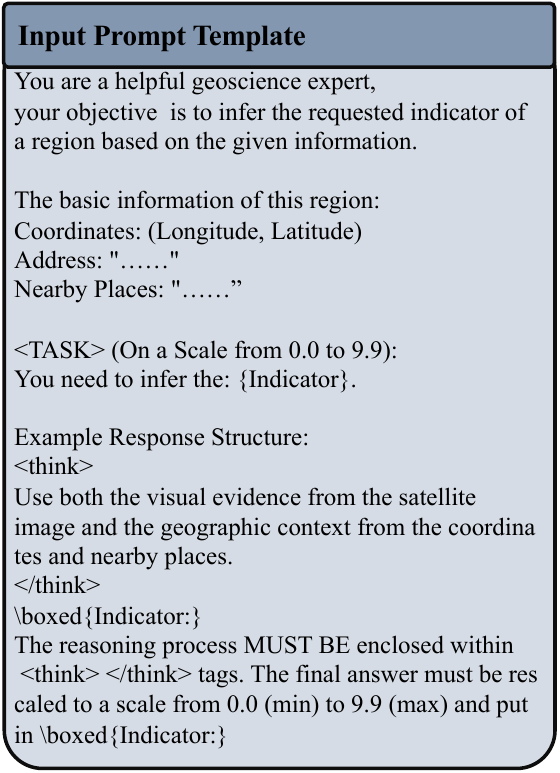}
    \label{fig:placeholder}
\end{figure}

\subsection{Dataset Details}\label{appendix_dataset}
We evaluate Urban-R1 on five urban indicators following GeoLLM~\cite{manvi2023geollm}: \emph{Population}, \emph{Carbon}, \emph{GDP}, \emph{Poverty}, and an additional \emph{House Price} indicator constructed from public housing datasets for New York~\cite{ny_housing}, Melbourne~\cite{melbourne_housing}, and Singapore~\cite{singapore_housing}. These five indicators constitute the Urban Region Profiling (URP) dataset, which features stratified training/validation/test splits; summary statistics are provided in Table~\ref{tab:urp_dataset}.

Five downstream urban tasks are constructed using diverse modalities (Table~\ref{tab:downstream_tasks}):
\begin{itemize}[leftmargin=*,topsep=0em,itemsep=0em]
    \item Site Selection: Built using coordinates entirely non-overlapping with the URP dataset, focusing on judging the suitability of locations for specific commercial establishments based on geographic text information and satellite imagery.
    \item Scene Function: Uses two satellite images with coordinates that do not overlap with the URP dataset and requires models to select the image containing the largest number of specified POIs, such as restaurants, bakeries.
    \item Land Use: Sampled from the open-source Aerial Image Dataset (AID) \cite{xia2017aid}, a benchmark dataset for aerial scene classification covering 30 land use types.
    \item Geo-localization: Adopts coordinates consistent with the Site Selection task, and generates three negative candidate coordinates (1,000 km away from the real coordinate) randomly to form a 4-option matching task.
    \item Urban Perception: Sampled from the open-source dataset associated with the study \cite{dubey2016deep}, focusing on perceptual judgments of streetscape attributes like livability and safety.
\end{itemize}

\begin{table}[!h]
\centering
\begingroup
\setlength{\tabcolsep}{2pt} 
\begin{tabular}{l|cccc} 
\toprule
\textbf{Indicator} & \textbf{Train} & \textbf{Val} & \textbf{Test (Seen)} & \textbf{Test (Unseen)} \\
\hline
GDP         & 1270 & 376 & 507 & 284 \\
Carbon      & 1235 & 372 & 501 & 284 \\
Population  & 1261 & 372 & 502 & 284 \\
Poverty     & 1234 & 370 & 502 & 231 \\
House Price & 1000 & 310 & 388 & 226 \\
\bottomrule
\end{tabular}
\endgroup
\caption{Statistics of the Urban Region Profiling dataset for different indicators and data splits.}
\label{tab:urp_dataset}
\end{table}

\begin{table}
\centering
\normalsize
\begin{tabular}{l|l|l}
\toprule
\textbf{Task Name} & \textbf{Size} & \textbf{Input Modality} \\ \hline
Site Selection & 550 & SAT + Text Info \\ 
Scene Function & 1000 & SAT \\ 
Land Use& 500 & SAT \\ 
Geo-localization & 1000 & SAT \\ 
Urban Perception & 800 & Streetview \\ 
\bottomrule
\end{tabular}
\caption{Details of downstream tasks. SAT: Satellite Imagery; Text Info: Location and geographic information; Streetview: Streetview Imagery.}
\label{tab:downstream_tasks}
\end{table}

\subsection{Implementation Details}
\label{app:implementation}
The SFT baselines are trained for 10 epochs (or steps, as specified in the main text) using model-specific pipelines. Specifically, the Qwen-family models (Qwen2.5-VL-3B/7B) are fine-tuned with LLaMA Factory \cite{llamafactory}, an open-source framework supporting efficient supervised fine-tuning of large language and multimodal models. The InternVL models (InternVL2.5-4B) are trained using the official InternVL training pipeline \cite{internvl}, which provides optimized configurations for vision-language pretraining and instruction tuning.


\end{document}